\documentclass[runningheads]{llncs}
\usepackage[T1]{fontenc}
\usepackage{array}
\usepackage{amsmath}
\usepackage{multirow}
\usepackage{booktabs}
\usepackage{listings}
\usepackage{makecell}
\usepackage{tcolorbox}
\usepackage{graphicx}
\usepackage{tikz}
\usepackage{url}
\usetikzlibrary{positioning,fit,calc}
\usepackage[ruled,vlined]{algorithm2e}
\begin{document}
\title{The Pragmatic Persona: Discovering LLM Persona through Bridging Inference}
\titlerunning{The Pragmatic Persona}
%
\author{Jisoo Yang\inst{1}\thanks{These authors contributed equally}\orcidID{0009-0009-7692-8482} \and 
Jongwon Ryu\inst{1}\textsuperscript{$\star$}\orcidID{0009-0003-9527-0774}\and \\
Minuk Ma\inst{2}\orcidID{0000-0003-0416-8479}\and
Trung X. Pham\inst{3}\orcidID{0000-0003-4177-7054}\and
Junyeong Kim\inst{1}\orcidID{0000-0002-7871-9627}}
\authorrunning{Yang et al.}

\institute{Department of Artificial Intelligence, Chung-Ang University, \\Seoul, 06974, Republic of Korea \\ \email{\{yjs229, fbwhddnjs511, junyeongkim\}@cau.ac.kr} \and Department of Computer Science, University of British Columbia, \\ Vancouver, BC V6T 1Z4, Canada \\ \email{minukma@cs.ubc.ca} \and Van Lang University, Ho Chi Minh City, Vietnam \\ \email{trung.px@vlu.edu.vn}\\
}

%
\maketitle             
\begin{abstract}
Large Language Models (LLMs) reveal inherent and distinctive personas through dialogue.
However, most existing persona discovery approaches rely on surface-level lexical or stylistic cues, treating dialogue as a flat sequence of tokens and failing to capture the deeper discourse-level structures that sustain persona consistency.
To address this limitation, we propose a novel analytical framework that interprets LLM dialogue through bridging inference—implicit conceptual relations that connect utterances via shared world knowledge and discourse coherence. By modeling these relations as structured knowledge graphs, our approach captures latent semantic links that govern how LLMs organize meaning across turns, enabling persona discovery at the level of discourse coherence rather than surface realizations.
Experimental results across multiple reasoning backbones and target LLMs, ranging from small-scale models to 80B-parameter systems, demonstrate that bridging-inference graphs yield significantly stronger semantic coherence and more stable persona identification than frequency or style-based baselines. These results show that persona traits are consistently encoded in the structural organization of discourse rather than isolated lexical patterns.
This work presents a systematic framework for probing, extracting, and visualizing latent LLM personas through the lens of Cognitive Discourse Theory, bridging computational linguistics, cognitive semantics, and persona reasoning in large language models. Codes are available at \url{https://github.com/JiSoo-Yang/Persona_Bridging.git}
\end{abstract}

\section{Introduction}

Large Language Models (LLMs) often show distinct and consistent behavioral patterns-sometimes described as their \textit{persona}. 
These personas appear naturally through dialogue, reflecting the model’s internal preferences and reasoning style. 
However, most existing studies identify personas using surface-level features such as word choice, tone, or prompt-based labels. 
Such methods fail to explain how LLMs maintain semantic coherence or why certain traits consistently emerge during interaction.
Human conversation, in contrast, depends not only on explicit language but also on implicit meaning. 
When people talk, they fill in gaps between ideas using background knowledge, a process known as \textit{bridging inference} \cite{asher1998bridging,clark1977comprehension,irmer2011bridging}. 
This process links one utterance to another by reasoning about part–whole, cause–effect, or thematic relations. 
It enables discourse to remain coherent even when key information is unstated.

\begin{figure}[t]
    \centering
    \includegraphics[width=\linewidth]{figure/figure1.pdf}
    \caption{
Example comparison of persona inference with and without bridging inference.
Without bridging, the model yields a surface-level description (\textit{photographer}),
whereas incorporating bridging relations enables deeper reasoning that captures reflective and emotional traits (\textit{traveler}).
This illustrates how bridging inference refines persona understanding beyond lexical cues.
}
    \label{figure1}
\end{figure}

Inspired by this cognitive mechanism, we propose a new framework that interprets LLM dialogues through the lens of bridging inference. As illustrated by the comparative example in Fig.~\ref{figure1}, our approach enables the construction of bridging-inference graphs that represent implicit conceptual links between utterances, allowing us to visualize how an LLM connects ideas and maintains semantic consistency. Specifically, our framework operationalizes seven types of bridging relations defined by Irmer \cite{irmer2011bridging}, following the taxonomy detailed in Table~\ref{tab:bridging_taxonomy}, to reveal the deeper conceptual structures that underlie an LLM’s persona expression.

Our main contributions are as follows:
\begin{enumerate}
    \item We propose a cognitively grounded framework for LLM persona analysis based on bridging inference, which yields more semantically coherent and interpretable persona representations than traditional lexical approaches.
    \item We implement predefined seven canonical bridging relation\cite{irmer2011bridging} types as few-shot exemplars for LLM reasoning analysis.
\end{enumerate}

Overall, this work bridges cognitive discourse theory and computational modeling, offering a new way to understand how LLMs build meaning and express consistent personas through dialogue.

\section{Related Work}

Research on Large Language Model (LLM) personas has transitioned from basic role-playing scenarios toward the sophisticated identification and steering of latent characteristics. We categorize the existing literature into three core domains: (1) the evaluation of predefined persona assignment and consistency, (2) the active inference of latent identities from generated responses, and (3) structural approaches to discourse modeling. This section situates our work within these domains, highlighting the shift from surface-level lexical analysis to deeper structural reasoning.

\subsection{Persona Assignment and Consistency Evaluation}

The field of LLM personas is broadly categorized into role-playing, where models adopt specific identities, and personalization, where they reflect user characteristics \cite{tseng2024survey}. Within this landscape, a significant portion of research has focused on persona assignment, where specific roles or traits are injected into the model via prompting to evaluate the similarity between the model's responses and the assigned identity \cite{jiang2023evaluating,li2024steerability}. This paradigm is further extended to social simulation environments, as exemplified by \textit{Generative Agents} \cite{park2023generative}, which are designed to mimic human-like interaction patterns based on profiles detailed in the prompts. Most existing studies in this domain primarily aim to verify the model’s ability to maintain these predefined identities, utilizing psychometric metrics like the Big Five personality traits to evaluate the alignment and consistency between the model’s responses and the assigned persona \cite{jiang2024personalllm}.

\subsection{Persona Inference}

Beyond simple behavioral steering via assigned prompts, recent studies have begun to explore persona inference, aiming to identify a model’s underlying identity from its generated responses. Data-driven frameworks extract latent personas from observed response distributions \cite{li2024steerability}, shifting the focus from explicit persona assignment to uncovering inherent behavioral patterns encoded in LLMs. This line of research is closely related to model alignment and safety, as actively inferring a model’s persona enables the prediction and control of its latent preferences.
Frameworks such as \textit{LLM-Mirror} \cite{kim2024llmmirror} leverage persona simulations to reproduce human-like response patterns for survey-style analysis, while related work evaluates inferred personas using psychometric constructs such as the Big Five personality traits \cite{jiang2024personalllm}. However, the controllability of these latent traits remains limited: quantitative analyses of prompt steerability show that certain persona dimensions are deeply rooted in pre-training data and resistant to surface-level control \cite{miehling2025promptsteerability}. Consequently, recent work has emphasized the necessity of active persona inference prior to response generation, particularly for aligning smaller language models with specific preferences \cite{tang2025activepersona}. Despite these advances, existing persona inference approaches largely rely on surface-level lexical cues and fail to capture the deeper logical and structural connections that emerge within discourse.

\subsection{Structural and Graph-based Approaches}
To address the limitations of surface-level lexical analysis, an emerging line of research utilizes structured representations to model dialogue and persona. Recent works have introduced persona-aware knowledge graphs (KGs) to maintain coherence in multi-turn or multi-session conversations \cite{multiturnsummarization2025,hu2025efficient,wang2024ai}. For example, constructing speaker-specific persona graphs and processing them via Graph Neural Networks (GNNs) has shown promise in enhancing dialogue summarization and long-term memory \cite{multiturnsummarization2025}. Similarly, Hu et al. \cite{hu2025efficient} propose a compact knowledge-grounded model that utilizes distilled knowledge to ensure role-playing consistency.
While these approaches leverage explicit knowledge structures, they often rely on predefined speaker labels or external knowledge bases, which may not capture the fluid nature of human discourse. Discourse theory suggests that coherence is maintained through implicit links—a concept rooted in the "given-new contract" \cite{clark1977comprehension} and formal semantics of bridging \cite{asher1998bridging,irmer2011bridging}. Although computational methods for bridging anaphora resolution have been studied \cite{gardent2003bridging,hou2013global}, they have rarely been applied to the problem of persona discovery. Our study distinguishes itself by focusing on these implicit conceptual patterns. Unlike existing graph-based methods that model explicit entities, our framework constructs discourse graphs from latent semantic links inferred through bridging, offering a deeper structural perspective on how LLMs maintain a coherent persona through cognitive reasoning rather than just pattern matching.


\section{Bridging Inference}


\begin{table}[!t]
  \centering
  \caption{Seven categories of bridging inference relations based on Irmer \cite{irmer2011bridging}. 
  Each relation type connects a previously mentioned entity (anchor) and a newly introduced referent (bridge).}
  \label{tab:bridging_taxonomy}
  \scriptsize
  \renewcommand{\arraystretch}{1.2}

  \begin{tabular}{p{2.56cm} | p{2.08cm} | p{4.44cm} | p{2.80cm}}
    \toprule
    \multicolumn{1}{c|}{\textbf{Relation Class}} &
    \multicolumn{1}{c|}{\textbf{Relation Type}} &
    \multicolumn{1}{c|}{\textbf{Description}} &
    \multicolumn{1}{c}{\textbf{Example}} \\
    \midrule

    \multirow{2}{*}{\parbox{2.56cm}{\raggedright\textit{Mereological Relations}}} &
    part-of &
    A mereological relation representing a physical or conceptual component of a larger whole. &
    \textit{engine $\rightarrow$ car} \\

    & member-of &
    Indicates an element belonging to a specific collective, group, or set. &
    \textit{student $\rightarrow$ class} \\

    \midrule

    \multirow{5}{*}{\parbox{2.56cm}{\raggedright\textit{Frame-related Relations}}} &
    instrument &
    Describes a tool or means fundamentally required to perform a specific action frame. &
    \textit{knife $\rightarrow$ cutting} \\

    & theme &
    Denotes an entity that is central to or affected by an eventuality. &
    \textit{topic $\rightarrow$ discussion} \\

    & cause-of &
    Establishes a causal or resultative dependency between two eventualities. &
    \textit{effort $\rightarrow$ success} \\

    & in &
    Specifies a relationship of spatial containment or situational inclusion. &
    \textit{book $\rightarrow$ library} \\

    & temporal &
    Captures sequential or simultaneous co-occurrence in time. &
    \textit{morning $\rightarrow$ breakfast} \\

    \bottomrule
  \end{tabular}
\end{table}

\subsection{Definition and Cognitive Foundation}
Bridging inference is defined as the cognitive process of establishing discourse coherence\cite{asher2003logics,murphy1985processes,huang2000discourse} by linking a newly introduced entity to a previously mentioned ``eventuality'' (events, states, or processes) through implicit semantic relations \cite{clark1977comprehension,irmer2011bridging,asher1998semantics,baker1998berkeley,bos1995bridging}. Building on Segmented Discourse Representation Theory (SDRT) \cite{asher2003logics} and Frame Semantics \cite{fillmore1976frame,malt1985role}, we interpret bridging not merely as a linguistic repair mechanism, but as a fundamental byproduct of maximizing discourse coherence (MDC)\cite{chierchia2009dynamics,johnson1977thinking,cohen2007anaphora}.

According to Irmer \cite{irmer2011bridging,fillmore2003background}, every eventuality introduced in a discourse evokes a corresponding semantic frame. This frame automatically activates Weak Discourse Referents---abstract entities that are not explicitly mentioned but are conceptually evoked as necessary participants in a situation. For instance, the lexical unit `murder' evokes a Killing frame, which inherently includes roles for a \textit{killer}, a \textit{victim}, and an \textit{instrument}. While these roles often remain unstated in surface text, they exist as underspecified variables in the speaker's mental representation of the discourse\cite{farkas2003semantics,kamp2013discourse,koenig1999definites}.

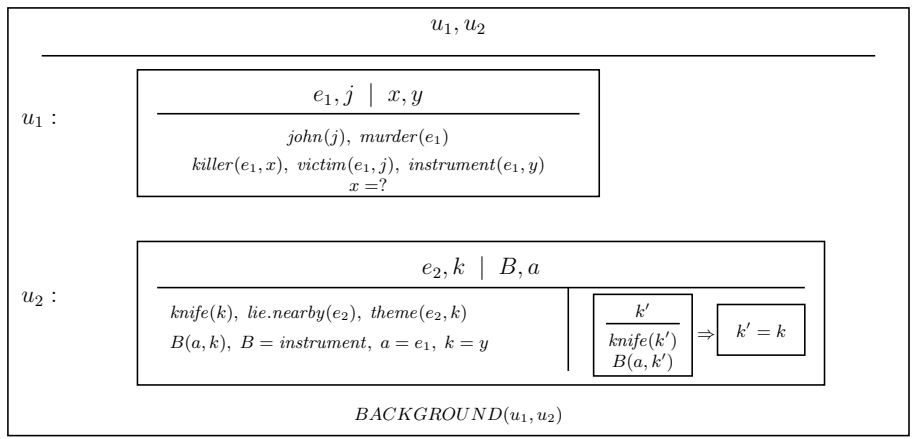
\begin{figure}[t]
\centering
\resizebox{12cm}{!}{%
\begin{tikzpicture}[font=\small, line width=0.8pt]

\def\W{16.0}
\def\H{7.6}
\draw (0,0) rectangle (\W,\H);

\node[font=\bfseries\large] at (\W/2,\H-0.35) {$u_1, u_2$};
\def\RuleY{\H-0.85}
\draw (0.6,\RuleY) -- (\W-0.6,\RuleY);

\def\TopGap{0.25}
\def\uOneTop{\RuleY-\TopGap}
\def\uTwoGap{0.80}

\def\uOneX{2.3}
\def\uOneW{8.2}
\def\uOneH{2.25}
\def\uOneY{\uOneTop-\uOneH}

\node[anchor=east, font=\large] at (1.0,\uOneY+0.60*\uOneH) {$u_1:$};
\draw (\uOneX,\uOneY) rectangle ++(\uOneW,\uOneH);

\node[font=\bfseries\large] at (\uOneX+\uOneW/2,\uOneY+\uOneH-0.45) {$e_1, j \;\mid\; x, y$};
\draw (\uOneX+0.35,\uOneY+\uOneH-0.78) -- (\uOneX+\uOneW-0.35,\uOneY+\uOneH-0.78);

\node[align=center] at (\uOneX+\uOneW/2,\uOneY+1.05)
  {$\mathit{john}(j),\; \mathit{murder}(e_1)$};
\node[align=center] at (\uOneX+\uOneW/2,\uOneY+0.55)
  {$\mathit{killer}(e_1,x),\; \mathit{victim}(e_1,j),\; \mathit{instrument}(e_1,y)$};
\node[align=center] at (\uOneX+\uOneW/2,\uOneY+0.20) {$x=?$};

\def\uTwoX{2.3}
\def\uTwoW{12.2}
\def\uTwoH{2.55}
\def\uTwoTop{\uOneY-\uTwoGap}
\def\uTwoY{\uTwoTop-\uTwoH}

\node[anchor=east, font=\large] at (1.0,\uTwoY+0.60*\uTwoH) {$u_2:$};
\draw (\uTwoX,\uTwoY) rectangle ++(\uTwoW,\uTwoH);

\node[font=\bfseries\large] at (\uTwoX+\uTwoW/2,\uTwoY+\uTwoH-0.48) {$e_2, k \;\mid\; B, a$};
\def\uTwoRuleY{\uTwoY+\uTwoH-0.82} 
\draw (\uTwoX+0.35,\uTwoRuleY) -- (\uTwoX+\uTwoW-0.35,\uTwoRuleY);

\def\RightMargin{0.35}              
\def\kAreaW{4.55}                   

\draw (\uTwoX+\uTwoW-\kAreaW,\uTwoRuleY) -- (\uTwoX+\uTwoW-\kAreaW,\uTwoY+0.25);

\node[anchor=north west, align=left, text width=7.4cm]
  at (\uTwoX+0.45,\uTwoRuleY-0.18) {%
$\mathit{knife}(k),\; \mathit{lie.nearby}(e_2),\; \mathit{theme}(e_2,k)$\\[4pt]
$B(a,k),\; B=\mathit{instrument},\; a=e_1,\; k=y$
};

\def\kBoxW{1.75}
\def\kBoxH{1.45}

\coordinate (kTL) at (\uTwoX+\uTwoW-\kAreaW+0.45,\uTwoRuleY-0.12);

\draw (kTL) rectangle ++(\kBoxW,-\kBoxH);

\node at ($(kTL)+(\kBoxW/2,-0.28)$) {$k'$};
\draw ($(kTL)+(0.22,-0.52)$) -- ($(kTL)+(\kBoxW-0.22,-0.52)$);
\node[align=center] at ($(kTL)+(\kBoxW/2,-1.02)$) {$\mathit{knife}(k')$\\ $B(a,k')$};

\node at (\uTwoX+\uTwoW-\kAreaW+2.45,\uTwoRuleY-0.85) {$\Rightarrow$};

\def\eqW{1.55}
\def\eqH{0.90}

\coordinate (eqTR) at (\uTwoX+\uTwoW-\RightMargin,\uTwoRuleY-0.30);

\draw (eqTR) rectangle ++(-\eqW,-\eqH);
\node at ($(eqTR)+(-\eqW/2,-\eqH/2)$) {$k'=k$};

\node[font=\small\bfseries] at (\W/2,\uTwoY-0.55) {$BACKGROUND(u_1,u_2)$};

\end{tikzpicture}%
}
\caption{SDRS-based resolution of a bridging inference (adapted from Irmer~\cite{irmer2011bridging}). The diagram illustrates the logical unification ($k=y$) where the newly introduced entity $k$ is anchored to the underspecified instrument slot $y$ evoked by the preceding event $e_1$}
\label{figure2_new}
\end{figure}

Fig.~\ref{figure2_new} illustrates this process through a Segmented Discourse Representation Structure (SDRS) based on the neo-Davidsonian event semantics \cite{parsons1990events,webber2016formal}. In the first utterance $u_1$ (``John was murdered''), the event $e_1$ introduces underspecified variables such as the killer ($x$) and the instrument ($y$), which are inherent to the frame but remain unassigned ($x=?$). When the subsequent utterance $u_2$ (``The knife lay nearby'') introduces a new entity, `the knife' ($k$), the system seeks to maximize coherence by identifying a bridging relation $B$. As shown in the diagram, the relation $B(a, k)$ is resolved as an \textit{instrument} link, anchoring the knife ($k$) to the event $e_1$. This results in a logical unification $k = y$, effectively plugging the newly introduced entity into the implicit slot of the previous event’s conceptual structure\cite{pustejovsky1991generative}.
In our framework, we posit that an LLM’s persona is embedded in how it resolves these weak referents. A model conditioned with a specific persona will consistently favor certain bridges over others based on its internal knowledge organization. For instance, the way an LLM connects a \textit{knife} to a \textit{murder} event through an instrument relation reflects a specific reasoning pattern that constitutes its latent persona. By capturing these implicit links, the PD-Agent can reconstruct the model's conceptual map which remains invisible in surface-level lexical analysis.

\subsection{Taxonomy of Bridging Relations}
To systematically categorize these inferences, we operationalize a taxonomy of seven bridging relations, broadly classified into two primary modes as proposed by Irmer~\cite{irmer2011bridging}: Association and Characterization. Instead of treating all semantic links equally, this distinction allows the PD-Agent to capture different layers of an LLM's conceptual organization.

The first mode, Indirect Reference by Association, encompasses mereological and structural dependencies. These relations (e.g., \textit{part-of}, \textit{member-of}) represent the model's ontological knowledge—how it perceives the hierarchical composition of the world. For instance, a persona centered around a \textit{systematic} or \textit{technical} role may exhibit a higher density of these structural links as it breaks down complex concepts into their constituent parts.

The second mode, Indirect Reference by Characterization, captures the dynamic and thematic roles within an evoked eventuality's frame. These relations (e.g., \textit{instrument}, \textit{theme}, \textit{cause-of}, \textit{in}, \textit{temporal}) reflect the model's situational reasoning, how it predicts the participants, causes, and spatio-temporal contexts of an action. A persona that is more narrative or empathy-driven might prioritize these characterization links to build a more vivid and contextually rich discourse.

\begin{figure}[t]
    \centering
    \includegraphics[width=\linewidth]{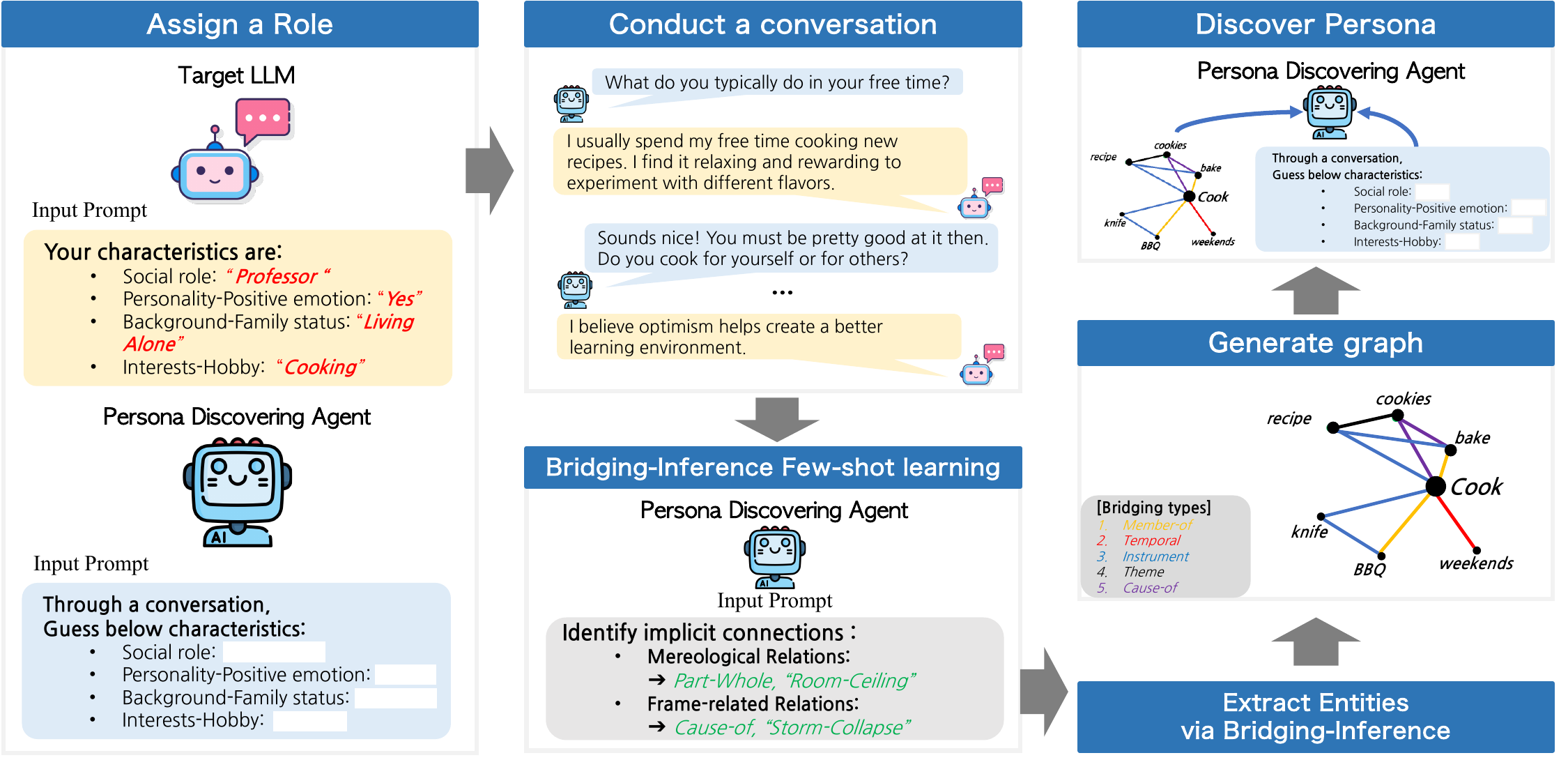}
    \caption{
  Overview of the \textbf{Persona-Discovering Agent (PD-Agent)} framework.  
  A hidden persona is first assigned to the Target LLM, followed by an adaptive dialogue that reveals implicit conceptual links through bridging inference (Table~\ref{tab:bridging_taxonomy}).  
  These relations are organized into a semantic graph from which the PD-Agent infers interpretable persona attributes across four dimensions.
  }
    \label{figure3}
\end{figure}

As summarized in Table~\ref{tab:bridging_taxonomy}, these seven categories serve as the edge labels for our semantic graph $G=(V,E)$~\cite{hou2013global}, which can be observed in the \textit{Generate graph} stage of Fig.~\ref{figure3}. By transforming unstructured dialogue into this labeled graph, the PD-Agent can analyze the distribution and density of different reasoning modes, providing a quantitative basis for latent persona discovery.

\section{The PD-Agent Framework}

\begin{algorithm}[t]
\caption{PD-Agent Persona Discovery Process}
\label{alg:pd_agent_process}
\KwIn{Target LLM $M_T$, PD-Agent $M_{PD}$, Persona Schema $\mathcal{S}$}
\KwOut{Predicted Persona Attributes $\hat{p}$}

\tcp{Step 1: Persona Assignment}
$p \leftarrow \text{Sample persona from } \mathcal{S}$\;
Condition $M_T$ with hidden prompt $p$\;

\tcp{Step 2: Interactive Interview}
$D \leftarrow \emptyset$ \tcp*{Dialogue history}
\For{$i \leftarrow 1$ \KwTo $N_{turns}$}{
    $q_i \leftarrow M_{PD}.\text{generate\_question}(D)$\;
    $r_i \leftarrow M_T.\text{generate\_response}(q_i)$\;
    $D \leftarrow D \cup \{(q_i, r_i)\}$\;
}

\tcp{Step 3: Bridging Inference Extraction}
$\mathcal{R} \leftarrow \emptyset$ \tcp*{Set of bridging relations}
$\mathcal{R} \leftarrow M_{PD}.\text{extract\_bridging\_relations}(D, \text{Taxonomy})$\;

\tcp{Step 4: Graph Construction and Reasoning}
$G \leftarrow (V, E)$ where $V \in \text{concepts in } \mathcal{R}, E \in \text{relations in } \mathcal{R}$\;
\ForEach{$v_i \in V$}{
    $\text{Importance}(v_i) \leftarrow \frac{\deg(v_i)}{\max_{v_j \in V} \deg(v_j)}$\;
}
$\hat{p} \leftarrow M_{PD}.\text{infer\_persona}(G, \text{Centrality})$\;
\Return $\hat{p}$\;
\end{algorithm}

We propose the \textbf{Persona-Discovering Agent (PD-Agent)} framework, an interactive pipeline designed to uncover the latent persona of large language models through bridging-inference-based reasoning. As illustrated in Fig.~\ref{figure3}, the framework facilitates a structured dialogue between two specialized agents:

\begin{itemize}
\item \textbf{PD-Agent (Reasoning Agent):} A controllable meta-agent responsible for conducting interviews, extracting semantic relations based on cognitive discourse theory, and inferring the target's persona.
\item \textbf{Target LLM (Persona Model):} A dialogue model conditioned with a hidden persona prompt (comprising social role, personality, etc.) to exhibit consistent behavioral traits.
\end{itemize}

The discovery process is executed in three distinct phases—interviewing, relation extraction, and graph-based reasoning—which are detailed in the following subsections.

\subsection{Framework Overview}                 
The PD-Agent framework is designed as a modular pipeline that transforms unstructured conversational data into a structured representation for persona recognition. As illustrated in the high-level workflow of Fig.~\ref{figure3}, the process follows a sequential path from role assignment to final persona discovery. The formal execution logic of this entire pipeline is summarized in Algorithm~\ref{alg:pd_agent_process}. By utilizing bridging-inference-based mining, the system bypasses surface-level lexical noise and captures the underlying reasoning patterns that define a model's latent persona.

\subsection{Structured Persona Schema}
To ensure a systematic evaluation, we define a four-dimensional persona schema that covers both sociological and psychological attributes. This schema, summarized in Table~\ref{tab:persona_schema}, serves as the ground truth for persona assignment and prediction. By structuring the persona into \textit{Social Role}, \textit{Personality (Big-Five)}, \textit{Background}, and \textit{Interests}, we can evaluate the framework's ability to identify diverse traits ranging from professional expertise to personal preferences.

\begin{table}[!t]
\centering
\caption{The four-dimensional persona schema used for Target LLM conditioning and PD-Agent evaluation.}
\label{tab:persona_schema}
\small
\renewcommand{\arraystretch}{1.1}
\begin{tabular}{@{} l|p{3.4cm}|p{6.6cm} @{}}
\toprule
\multicolumn{1}{c|}{\textbf{Dimension}} &
\multicolumn{1}{c|}{\textbf{Subcategory}} &
\multicolumn{1}{c}{\textbf{Representative Values}} \\ \midrule
\textbf{Social Role} & Professional & Doctor, Lawyer, Professor, Accountant \\
& Technical Management & Software Engineer, Data Scientist, Product Manager \\
& Public Service & Civil Servant, Police Officer, Teacher \\ \hline
\textbf{Personality} & Big-Five Traits & Openness, Conscientiousness, Extroversion, Agreeableness, Neuroticism \\ \hline
\textbf{Background} & Education & High School, Bachelor's, Master's, Ph.D. \\
& Location & Urban, Rural \\
& Family Status & Single, Married, Living Alone \\ \hline
\textbf{Interests} & Hobbies & Reading, Traveling, Gaming \\
& Core Values & Creativity, Family, Integrity \\
& Comm. Style & Direct, Emotional \\ \bottomrule
\end{tabular}
\end{table}

\subsection{Interactive Interview \& Extraction}
This phase corresponds to the \textit{Conduct a conversation} and \textit{Bridging-Inference Few-shot learning} stages in Fig.~\ref{figure3}. The process begins with an adaptive interview where the PD-Agent generates 3--5 turns of dialogue focused on open-ended topics. This stage is crucial for eliciting persona-revealing cues without direct self-disclosure from the Target LLM.
Once the dialogue is recorded, the PD-Agent performs Bridging Inference Extraction. As shown in the `Few-shot learning' block of Fig.~\ref{figure3}, the agent is provided with few-shot exemplars of mereological and frame-related relations to identify implicit connections. Using the meta-prompting logic in Listing\ref{lst:bridging_prompt_template}, the PD-Agent identifies \textit{anchor-anaphor} pairs and extracts the specific relation type along with its rationale. This ensures that every extracted link is grounded in cognitive discourse theory rather than simple keyword matching.

\begin{lstlisting}[
  float,
  floatplacement=t,
  basicstyle=\ttfamily\scriptsize,
  breaklines=true,
  columns=fullflexible,
  frame=single,
  caption={Bridging extraction prompt template used by PD-Agent.},
  label={lst:bridging_prompt_template}
]
TASK
Given the conversation below, identify implicit bridging relations that require world knowledge or frame understanding.
Use the seven relation types: {part-of, member-of, instrument, theme, cause-of, in, temporal}.

RETURN
For each relation, provide:
- anchor: 1-3 word concept that appears earlier
- anaphor: 1-3 word concept that is implicitly linked
- relation_type: one of {part-of, member-of, instrument, theme, cause-of, in, temporal}
- explanation: one-sentence rationale
- sentence_context: minimal spans from the dialogue that support the inference

CONSTRAINTS
- Use short noun phrases (1-3 words) for anchor/anaphor.
- Pure coreference (e.g., "car ... it") is not bridging.
- Prefer canonical concepts (e.g., "oven" over "the big oven in my kitchen").

FORMAT
Return ONLY valid JSON with a top-level key "bridging_relations".
\end{lstlisting}

\subsection{Semantic Graph Construction and Reasoning}

The final stage of the framework involves the \textit{Generate graph} and
\textit{Discover Persona} stages, as illustrated in
Fig.~\ref{figure3}. The extracted relations are transformed into a
directed semantic graph $G = (V, E)$, where nodes $V$ denote discourse
concepts (e.g., \textit{cook}, \textit{recipe}, \textit{bake}) and edges
$E$ encode the labeled bridging relations.
To quantify the importance of specific concepts in the model’s
persona-driven reasoning, we compute the normalized degree
centrality for each node $v_i \in V$:

\begin{equation}
\mathrm{Importance}(v_i)
=
\frac{\deg(v_i)}{\max_{v_j \in V} \deg(v_j)}
\end{equation}

where $\deg(v_i)$ denotes the total degree of node $v_i$, defined as the
sum of its degree. As depicted in the graph
visualization in Fig.~\ref{figure3}, nodes with higher centrality scores
serve as primary conceptual hubs. Finally, the PD-Agent analyzes these
central clusters together with the distribution of relation types (e.g.,
the prevalence of \textit{instrument} versus \textit{causal} links) to
predict the latent persona attributes across the four schema dimensions.




\section{Experiments}

\subsection{Experimental Setup}
We evaluated the PD-Agent framework across a diverse set of reasoning backbones and target models to ensure the robustness of the persona discovery process. For the Reasoning Backbones, we employed six state-of-the-art models: two proprietary models (\textit{GPT-4o}, \textit{Claude 3.5 Sonnet}), two reasoning-optimized models (\textit{Gemini 1.5 Pro}, \textit{o1-mini}), and two high-performance open-source models (\textit{DeepSeek-V3}, \textit{Llama-3.1-70B}).
The Target LLMs were categorized into two groups based on their parameter scales:
(1) \textbf{Small Targets} (\textit{Qwen3-1.7B}, \textit{Llama-3.1-8B}, \textit{Gemini-2.5-Flash}), and
(2) \textbf{Large Targets} (\textit{Qwen3-30B}, \textit{Llama-3.1-70B}, \textit{Qwen3-80B}).
Each backbone conducted an interactive interview with every Target LLM to generate dialogue data conditioned on hidden personas sampled from the schema.
We compared three reasoning strategies:
(1) Vanilla: Direct persona attribute prediction from dialogue text.
(2) Frequency-Aware: Leveraging token frequency distributions for trait identification.
(3) PD-Agent (Ours): Utilizing bridging-inference extraction and semantic graph construction.
Performance was measured using cosine similarity between the predicted and ground-truth persona representations across four dimensions.
All experiments were performed using four \textit{NVIDIA RTX Pro 6000 MaxQ (Blackwell)} GPUs, each equipped with 96GB of VRAM.

\begin{table}[!t]
  \centering
  \caption{Comprehensive evaluation of persona discovery similarity across varying Target LLM scales. \textbf{Small Targets}: Qwen3-1.7B, Llama-3.1-8B, Gemini-F (\textit{Gemini-2.5-Flash}). \textbf{Large Targets}: Qwen3-30B, Llama-3.1-70B, Qwen3-80B. Backbone models represent the reasoning agents, while Target models represent the models assigned with hidden personas.}
  \label{tab:results_full}
  \resizebox{1.0\textwidth}{!}{%
  \begin{tabular}{l|l|ccc|ccc|c}
    \toprule
    & & \multicolumn{3}{c|}{\textbf{Small Target LLMs}} & \multicolumn{3}{c|}{\textbf{Large Target LLMs}} & \\
    \textbf{Backbone} & \textbf{Method} & \textbf{Qwen3-1.7B} & \textbf{Llama-8B} & \textbf{Gemini-F} & \textbf{Qwen3-30B} & \textbf{Llama-70B} & \textbf{Qwen3-80B} & \textbf{Avg.} \\
    \midrule
    \textbf{Claude 3.5} & Vanilla  & 0.78 & 0.74 & 0.76 & 0.81 & 0.77 & 0.79 & 0.77 \\
    \textbf{Sonnet} & + Freq-Aware & 0.80 & 0.76 & 0.78 & 0.83 & 0.79 & 0.81 & 0.80 \\
    & \textbf{+ PD-Agent (Ours)} & \textbf{0.91} & \textbf{0.88} & \textbf{0.89} & \textbf{0.94} & \textbf{0.92} & \textbf{0.93} & \textbf{0.91} \\
    \midrule
    \textbf{Gemini 1.5} & Vanilla  & 0.85 & 0.82 & 0.83 & 0.86 & 0.83 & 0.85 & 0.84 \\
    \textbf{Pro} & + Freq-Aware & 0.83 & 0.80 & 0.82 & 0.87 & 0.84 & 0.86 & 0.84 \\
    & \textbf{+ PD-Agent (Ours)} & \textbf{0.93} & \textbf{0.90} & \textbf{0.91} & \textbf{0.96} & \textbf{0.94} & \textbf{0.95} & \textbf{0.93} \\
    \midrule
    \textbf{DeepSeek} & Vanilla  & 0.83 & 0.79 & 0.81 & 0.85 & 0.82 & 0.84 & 0.82 \\
    \textbf{V3} & + Freq-Aware & 0.85 & 0.81 & 0.83 & 0.87 & 0.84 & 0.86 & 0.84 \\
    & \textbf{+ PD-Agent (Ours)} & \textbf{0.94} & \textbf{0.91} & \textbf{0.92} & \textbf{0.97} & \textbf{0.94} & \textbf{0.96} & \textbf{0.94} \\
    \midrule
    \textbf{Llama-3.1} & Vanilla  & 0.79 & 0.75 & 0.77 & 0.82 & 0.78 & 0.80 & 0.78 \\
    \textbf{70B-Inst.} & + Freq-Aware & 0.81 & 0.77 & 0.79 & 0.84 & 0.80 & 0.82 & 0.81 \\
    & \textbf{+ PD-Agent (Ours)} & \textbf{0.91} & \textbf{0.87} & \textbf{0.88} & \textbf{0.94} & \textbf{0.91} & \textbf{0.92} & \textbf{0.90} \\
    \midrule
    \textbf{GPT-4o} & Vanilla  & 0.82 & 0.78 & 0.80 & 0.88 & 0.85 & 0.87 & 0.83 \\
    & + Freq-Aware & 0.88 & 0.85 & 0.87 & 0.90 & 0.87 & 0.89 & 0.88 \\
    & \textbf{+ PD-Agent (Ours)} & \textbf{0.97} & \textbf{0.95} & \textbf{0.90} & \textbf{0.98} & \textbf{0.97} & \textbf{0.98} & \textbf{0.96} \\
    \midrule
    \textbf{o1-mini} & Vanilla  & 0.86 & 0.83 & 0.84 & 0.89 & 0.86 & 0.87 & 0.86 \\
    \textbf{(Reasoning)} & + Freq-Aware & 0.88 & 0.85 & 0.86 & 0.91 & 0.88 & 0.89 & 0.88 \\
    & \textbf{+ PD-Agent (Ours)} & \textbf{0.98} & \textbf{0.97} & \textbf{0.98} & \textbf{0.99} & \textbf{0.98} & \textbf{0.99} & \textbf{0.98} \\
    \bottomrule
  \end{tabular}%
  }
\end{table}

\subsection{Results}
The comprehensive results in Table~\ref{tab:results_full} reveal that the PD-Agent framework consistently outperforms the baseline models across all backbones and target scales. On average, our method achieved a similarity score of 0.90 to 0.98, significantly exceeding the Vanilla and Frequency-Aware baselines.
Notably, the \textit{o1-mini} backbone, which is specifically optimized for logical reasoning, achieved the highest performance, reaching a similarity of 0.99 on Large Targets. This suggests that the quality of persona discovery is highly correlated with the backbone's ability to perform complex inference. Furthermore, the PD-Agent yielded a significant improvement (up to +0.15) over frequency-based heuristics, proving that persona traits are encoded in the structural coherence of discourse rather than simple lexical choices.
This performance gap is more pronounced for larger target models, indicating that the proposed framework scales favorably with model size compared to surface-level baselines.

\subsection{Scalability and Error Analysis}
Beyond the average performance, we analyzed the impact of Target LLM scale on persona discovery.

\begin{itemize}
\item \textbf{Scalability across Model Sizes}: As shown in Table~\ref{tab:results_full}, the PD-Agent's performance actually increases as the Target LLM becomes larger. While Large Targets (30B--80B) generate more complex and subtle linguistic patterns that can mask persona cues for the Vanilla model, the PD-Agent effectively exploits the richer semantic density of these models. The denser bridging-inference graphs produced from Large Targets provide more salient nodes for centrality analysis, leading to more accurate attribute inference.

\item \textbf{Category-specific Sensitivity}: Despite high overall performance, most prediction errors occurred in the \textit{background} category. Errors were often linked to ambiguous cues; for example, a Target LLM mentioning ``I have moved many times'' might be misclassified as being ``abroad'' (Background: Location) when its ground truth was actually ``living alone'' (Background: Family Status).
\end{itemize}

Qualitative visualizations in Fig.~\ref{figure3} confirm that bridging-based reasoning yields more interpretable and robust persona representations compared to surface-level analysis. The high stability of inference (standard deviation $< 0.03$ across five runs) further validates the framework as a reliable diagnostic tool for LLM behavior analysis.

\section{Conclusion}

In this paper, we presented the PD-Agent, a novel framework for discovering the latent personas of Large Language Models (LLMs) through cognitively grounded bridging inference and graph-structured reasoning. By operationalizing implicit semantic relations rather than relying on surface-level lexical cues, our approach provides a structural lens to interpret how LLMs maintain consistent behavioral patterns throughout dialogue.
Extensive experiments involving six state-of-the-art reasoning backbones and multiple high-parameter target models (up to 80B) demonstrate that the PD-Agent significantly outperforms traditional frequency-based baselines, achieving up to 0.99 cosine similarity with ground-truth persona attributes. Our findings underscore that as LLMs grow in scale and complexity, their persona traits are more effectively captured through the structural coherence of discourse rather than simple word choices.
This study bridges the gap between cognitive discourse theory and computational persona modeling, establishing discourse-level relational reasoning as a powerful diagnostic tool for understanding model behavior. Future research will extend this framework to multilingual and multimodal contexts, exploring how diverse cultural values and perceptual cues further shape and reveal the complex latent personas of next-generation language models.



The PD-Agent framework, while effective, has several limitations as identified through experimental analysis:

\begin{itemize}
    \item \textbf{Category-specific Sensitivity}: The highest frequency of prediction errors occurred in the \textit{background} category. These errors often stemmed from ambiguous linguistic cues; for example, a Target LLM's mention of frequent relocation was sometimes misclassified as "\textit{living abroad}" (Location) when the ground truth was "\textit{living alone}" (Family Status).
    \item \textbf{Dependency on Backbone Reasoning}: The quality of persona discovery is highly correlated with the reasoning backbone's ability to perform complex inference. Models optimized for logical reasoning, such as o1-mini, achieved the highest similarity scores, suggesting that the framework's effectiveness may be limited when using models with weaker inference capabilities.
    \item \textbf{Fixed Taxonomy of Bridging Relations}: The framework currently operationalizes a specific taxonomy of seven canonical bridging relations based on Irmer [9]. While this provides a structured analytical lens, it may not capture the full diversity of implicit semantic links present in natural, fluid discourse.
    \item \textbf{Scalability and Semantic Density}: Although the framework performs well across scales, Large Targets (30B-80B) generate richer semantic density, which leads to more salient nodes for centrality analysis. Conversely, smaller target models may offer fewer cues, potentially affecting the robustness of the resulting bridging-inference graphs.
    \item \textbf{Modal and Linguistic Scope}: The current study is primarily focused on textual dialogue within a specific set of parameters. Future research is required to evaluate how the framework performs in multilingual and multimodal contexts, where diverse cultural values and perceptual cues (such as visual or auditory data) may further shape persona expression.
\end{itemize}

\begin{credits}
\subsubsection{\ackname}
This work was partly supported by the Technology development Program [RS-2025-25443237] funded by the Ministry of SMEs and Startups(MSS, Korea) and partly supported by the Institute of Information \& Communications Technology Planning \& Evaluation (IITP) grant funded by the Korea government (MSIT) [RS-2021-II211341, Artificial Intelligence Graduate School Program (Chung-Ang University)].

\subsubsection{LLM Usage.}

Large Language Models (LLMs) were utilized in this study in two limited and clearly defined roles: (1) as reasoning agents and target models during the persona discovery experiments, and (2) as a writing assistant for language polishing.

During the experiments, LLMs were employed strictly as experimental subjects or analytical agents. Specifically, the Target LLMs were conditioned with hidden persona prompts and interacted with the Persona-Discovering Agent (PD-Agent) to generate dialogue data. Separately, various LLMs served as reasoning backbones for the PD-Agent, responsible for generating interview questions, extracting bridging-inference relations, and performing graph-based persona inference. In all cases, the models were used solely for inference and analysis; no model parameters were trained, fine-tuned, or updated using the experimental data.

In addition, an LLM was used as a writing assistant to polish the language of the manuscript. This usage was limited to improving clarity, grammar, and overall readability of the text, without contributing new experimental ideas, altering the scientific content, or influencing the interpretation of results. All methodological design, experimental setup, analysis, and conclusions were determined entirely by the authors.

\subsubsection{\discintname}
The authors have no competing interests to declare that are relevant to the content of this article.
\end{credits}

%
%
%
\bibliographystyle{splncs04}
\bibliography{mybibliography}

\end{document}